\documentclass[journal,twoside,web]{ieeecolor}
\usepackage{tmi}
\usepackage{cite}
\usepackage{amsmath,amssymb,amsfonts}
\usepackage{algorithmic}
\usepackage{graphicx}
\usepackage{textcomp}
\usepackage{booktabs}
\usepackage{multirow}
\usepackage{color}
\def\BibTeX{{\rm B\kern-.05em{\sc i\kern-.025em b}\kern-.08em
    T\kern-.1667em\lower.7ex\hbox{E}\kern-.125emX}}
\graphicspath{ {./fig/} }
\begin{document}
\title{Shapley Values-enabled Progressive Pseudo Bag Augmentation for Whole-Slide Image Classification}
\author{Renao Yan, \IEEEmembership{Student Member, IEEE}, Qiehe Sun, Cheng Jin, \IEEEmembership{Student Member, IEEE}, Yiqing Liu,\\ Yonghong He, Tian Guan, and Hao Chen, \IEEEmembership{Senior Member, IEEE}
\thanks{This work was supported by National Natural Science Foundation of China (No. 62202403), Shenzhen Science and Technology Innovation Committee Funding (Project No. SGDX20210823103201011), and Hong Kong Innovation and Technology Fund (No. PRP/034/22FX).}
\thanks{Renao Yan, Qiehe Sun, Yiqing Liu, Yonghong He, and Tian Guan are with Shenzhen International Graduate School, Tsinghua Unversity, Beijing, China.}
\thanks{Renao Yan, Cheng Jin, and Hao Chen are with the Department of Computer Science and Engineering, the Hong Kong University of Science and Technology, Hong Kong, China. Hao Chen is also affiliated with the Department of Chemical and Biological Engineering and Division of Life Science, Hong Kong University of Science and Technology, Hong Kong, China.}
\thanks{Corresponding author: Hao Chen (e-mail:	jhc@cse.ust.hk).}}

\maketitle

\begin{abstract}
In computational pathology, whole-slide image (WSI) classification presents a formidable challenge due to its gigapixel resolution and limited fine-grained annotations. Multiple-instance learning (MIL) offers a weakly supervised solution, yet refining instance-level information from bag-level labels remains challenging. While most of the conventional MIL methods use attention scores to estimate instance importance scores (IIS) which contribute to the prediction of the slide labels, these often lead to skewed attention distributions and inaccuracies in identifying crucial instances. To address these issues, we propose a new approach inspired by cooperative game theory: employing Shapley values to assess each instance's contribution, thereby improving IIS estimation. The computation of the Shapley value is then accelerated using attention, meanwhile retaining the enhanced instance identification and prioritization. We further introduce a framework for the progressive assignment of pseudo bags based on estimated IIS, encouraging more balanced attention distributions in MIL models. Our extensive experiments on CAMELYON-16, BRACS, TCGA-LUNG, and TCGA-BRCA datasets show our method's superiority over existing state-of-the-art approaches, offering enhanced interpretability and class-wise insights. Our source
code is available at https://github.com/RenaoYan/PMIL.
\end{abstract}


\section{Introduction}\label{sec:introduction}
\IEEEPARstart{R}{ecent} advancements in digital pathology and artificial intelligence have significantly expanded the potential for analyzing whole-slide images (WSIs) in diagnostic contexts, prognostic evaluations, and various clinical tasks \cite{ba2022assessment, schapiro2017histocat, bera2019artificial, niazi2019digital, moen2019deep, 8756037, lin2022pdbl, yan2023unpaired}. A key area within this domain is WSI classification \cite{pinheiro2015image, bejnordi2017diagnostic, campanella2019clinical, lee2022derivation, pati2022hierarchical, li2023deeptree}, a fundamental and vital process distinguished by the gigapixel resolution of WSIs, setting it apart from typical natural image classification. The complex nature of the WSI classification task necessitates the adoption of specialized methodologies such as multi-instance learning (MIL) \cite{maron1997framework, feng2017deep, zhu2017deep, li2021multi, wang2022weakly}. 

The principle of MIL is that the presence of at least one positive instance within a bag classifies the entire bag as positive; otherwise, it is considered negative. Most of the current research in MIL builds on the essential idea of distilling more instance-level information from bag-level labels \cite{pmlr-v143-sharma21a, chen2022scaling, zhu2023accurate, lin2023}. In this paradigm, attention-based pooling \cite{ilse2018attention} stands out as a prominent technique, as the attention score $\alpha$ it generates for each instance in the bag naturally serves as a choice for estimating the contribution of each instance, referred to as the instance importance score (IIS). For example, attention scores play a crucial role in assisting MIL models in discerning significant instances to mitigate overfitting \cite{lu2021data,li2021dual,yufei2022bayes,zhang2023attention}. Moreover, some studies leverage attention scores to fine-tune the feature encoder \cite{qu2022bi,yu2023bayesian,li2023task}. The attention score is so powerful that these studies all operate under the assumption that crucial (positive) instances can be identified by selecting those with top-ranking attention scores.

\begin{figure*}[]
  \centering
  \setlength{\abovecaptionskip}{0pt}%
  \setlength{\belowcaptionskip}{0pt}%
  \includegraphics[width=\textwidth]{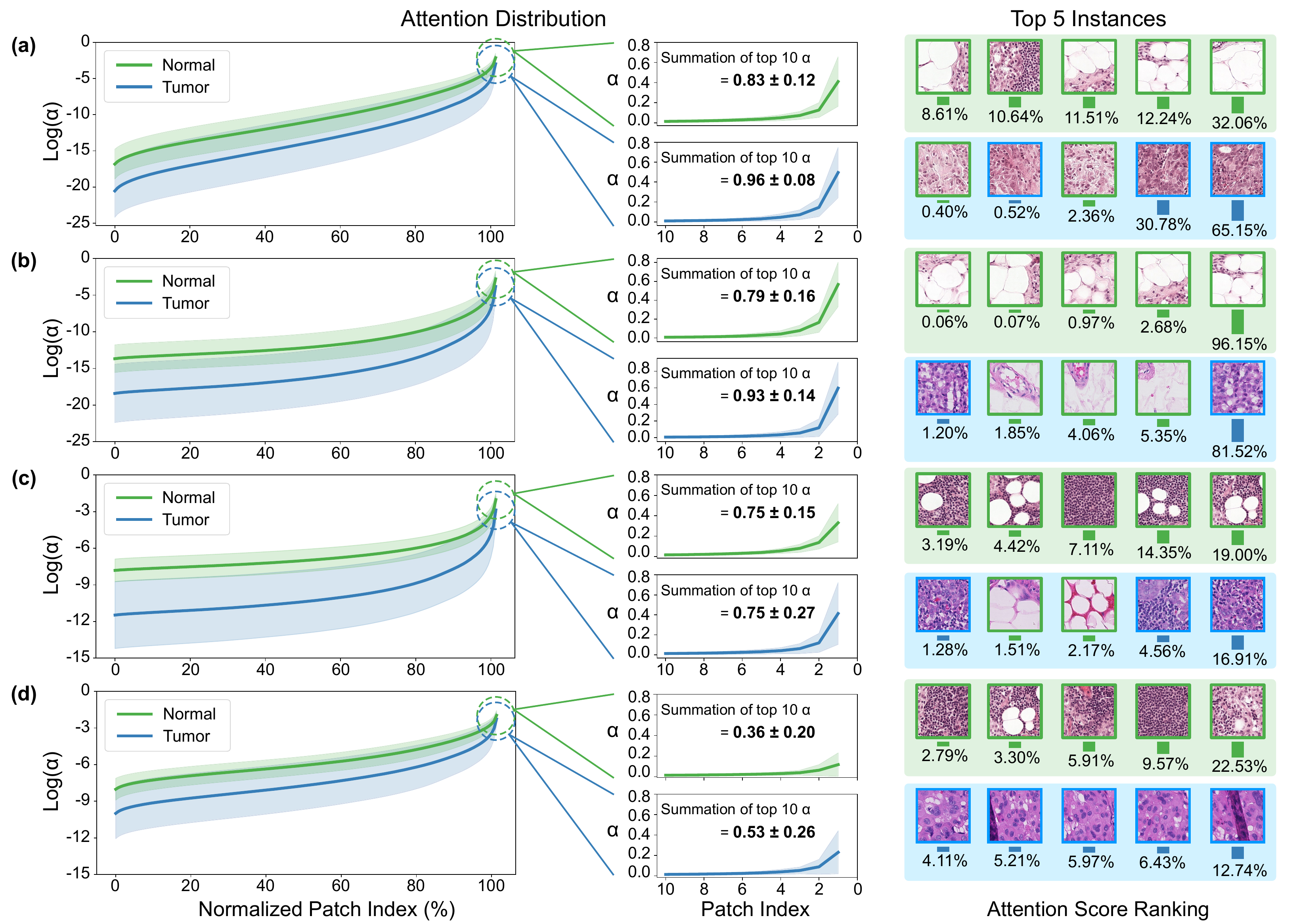}
  \caption{Observation of attention distributions and top 5 instances of one example slide in the CAMELYON-16 Dataset. (a)-(d) employs ABMIL, CLAM, DTFD, and proposed PMIL as the MIL model, respectively. In the column of "Attention Distribution", the patch index is normalized to a range of 0 to 1 for all patches across all slides in the left sub-figure. Notably, the distribution of attention scores is skewed, with a few instances accumulating a significant share. In the column of "Top 5 Instances", positive instances (depicted in green border) are not consistently ranked in order of attention scores, as negative instances (depicted in blue border) may take precedence in the queue.}\label{attn}
\end{figure*}

Nevertheless, our empirical investigations, as illustrated in Fig. \ref{attn}(a)-(c), expose the challenges encountered by attention-based MIL, namely: 
\begin{enumerate}
    \item \textbf{Extreme distribution of attention.} A limited number of instances receive the majority of attention scores. For example, the summation of the top 10 attention scores accounts for 75\% or more. This concentration can lead to insufficient training. 
    \item \textbf{Misidentification of positive instances via top-ranking attention scores.} Positive instances are not guaranteed to rank at the top. Both positive and negative instances can be filtered out using top-$k$ attention scores. Assigning these instances solely positive labels can introduce noise during training or fine-tuning.
\end{enumerate}

To address these two inherent problems within attention-based MIL, in this study, we propose a progressive pseudo bag augmented MIL framework, termed PMIL. This framework takes full advantage of pseudo bag augmentation under the guidance of the Shapley value. Derived from the cooperative game \cite{shapley1953value}, the Shapley value offers a solution by quantifying the contribution of each instance based on its interactions with others \cite{messalas2019model,tang2021data}. Specifically, we first apply pseudo bag augmentation to MIL, aiming at encouraging models to focus on more important instances. Furthermore, to improve the mislabeling issue in pseudo bag augmentation, we introduce the Shapley value as a means of IIS estimation to constrain the assignment strategy instead of random splitting. Our approach divides a regular bag into a series of pseudo bags in a reasonable manner, thereby reducing the intrinsic noise associated with pseudo bag creation and enhancing the model's generalization ability. In summary, our main contributions are as follows:

\begin{itemize}
  \item Acknowledging the limitations of attention score-based IIS in terms of ranking accuracy and interpretation, we introduce an accelerated Shapley value with linear computational complexity to measure IIS in the context of multiple-instance learning for the first time.
  \item With Shapley value-based IIS, we propose a progressive pseudo bag augmented multiple-instance learning framework, effectively bolstering MIL performance.
  \item Extensive experiments on the CAMELYON-16, BRACS, TCGA-LUNG, and TCGA-BRCA datasets demonstrate that our method outperformed other state-of-the-art methods in both slide-level and instance-level evaluation, and provided class-wise interpretation with Shapley values.
\end{itemize}

\section{Related Work}
\subsection{Multiple Instance Learning for WSI Classification}
In the field of whole-slide image classification, due to the gigapixel resolution and lack of manual annotation, multiple-instance learning serves as a promising weakly supervised learning approach. Lu \textit{et al.} \cite{lu2021data} introduces an additional cluster branch (CLAM) founded on IIS estimated by attention scores to distinguish features via projection. Wang \textit{et al.} \cite{wang2022scl} apply contrastive learning to facilitate the interaction of intra-WSI and inter-WSI information using attention scores to filter positive, negative, and hard negative instances. Shao \textit{et al.} \cite{shao2021transmil} proposed Transformer-based MIL (TransMIL), introducing the self-attention mechanism to express the correlated relationship between instances. Yu \textit{et al.} \cite{yu2023bayesian} assigns slide-level labels to patches garnering the highest attention to fine-tune the feature encoder through an auxiliary patch classification task. Li \textit{et al.} \cite{li2023task} selected instances with top-ranking attention scores for end-to-end MIL training, addressing the information bottleneck. Zhang \textit{et al.} \cite{zhang2023attention} introduce multiple branch attention to capture more discriminative instances, and stochastic top-$k$ instance masking to suppress salient instances.

\begin{figure*}[]
  \centering
  \setlength{\abovecaptionskip}{0pt}%
  \setlength{\belowcaptionskip}{0pt}%
  \includegraphics[width=\textwidth]{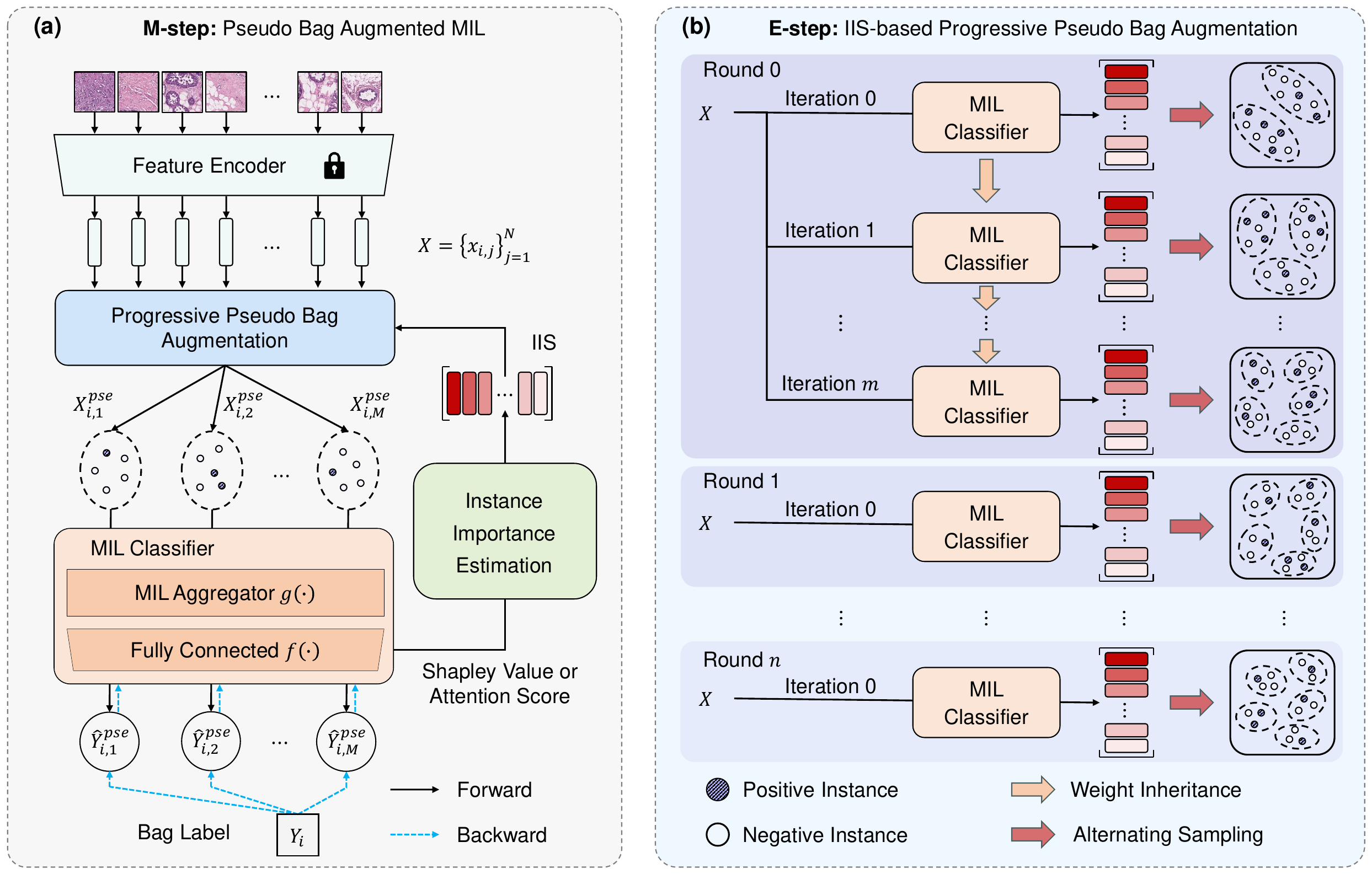}
  \caption{Overview of the proposed PMIL framework. (a) A collection of patches, extracted from a WSI, is partitioned into $M$ ($M$ gradually increases) pseudo bags based on their estimated IIS, and then are trained in the same manner as regular bags. (b) The weights of the MIL model are frozen to estimate IIS, facilitating pseudo bag assignment. The number $M$ of pseudo bags progressively increases at the iteration when the MIL model converges in round 0, and initial pseudo bags are assigned using IIS estimated by the MIL model in the previous round. \textbf{Note that the pseudo bag augmentation is only used during the training process.}}\label{overview}
\end{figure*}
\subsection{Advanced IIS related techniques for MIL}
Addressing the challenge of extreme attention distribution, a notable strategy involves dividing the regular bag into several pseudo bags \cite{shao2021weakly, yang2023protodiv}. This approach encourages MIL models to learn from a greater variety of bags. Existing pseudo bag-based approaches adopted a random splitting strategy. For instance, DTFD \cite{zhang2022dtfd} proposed a double-tier MIL model upon features distilled from pseudo bags to solve the mislabeling issue. As depicted in Fig. \ref{attn}(c), DTFD smoothed out the attention distribution compared to other non-pseudo-bag-augmented MIL methods, which verifies the efficacy of pseudo bag augmentation to some extent. PseMix \cite{liu2024pseudo} applied a bag-prototype-based clustering to constrain the random pseudo bag assignment, and then introduced the MixUp strategy to assign mixed pseudo bags with soft targets for training.

In the context of the misidentification of positive instances, Javed \textit{et al.} \cite{javed2022additive} applied the Shapley value as an alternative to attention scores for MIL training and inference, considering it as the weighting score of each feature.

In summary, current pseudo bag augmented MIL methods typically employ a random splitting strategy, which inevitably introduces mislabeling noise into both training and inference phases. Additionally, their pseudo bag assignment is static and cannot be learned during training. Although Javed et al. \cite{javed2022additive} were the first to introduce the Shapley value to MIL, they did not address the significant computational demands of calculating Shapley values, which presents a major challenge for practical applications. Furthermore, their code lacks detailed training procedures, making it difficult to reproduce their performance results.

\section{Method}
In this section, we will first retrospect the MIL paradigm and pseudo bag augmentation technique, then introduce Shapley value-based IIS to improve pseudo bag assignment, and finally propose our framework, as shown in Fig. \ref{overview}.

\subsection{Problem and Notation}
\subsubsection{MIL in WSI classification} In this task, the training set of labeled WSIs is denoted as $\mathcal{D}=\left\{X_i,Y_i\right\}_{i=1}^{\left|\mathcal{D}\right|}$, where $X_i=\left\{x_{i,j}\right\}_{j=1}^{N_i}$ represents the $i$-th bag (slide) comprising $N_i$ instances after feature extraction. Our objective is to learn the mapping: $\mathcal{X}\rightarrow \mathcal{Y}$, where $\mathcal{X}$ is the bag space, and $\mathcal{Y}$ is the label space. The conventional MIL classifier maps the aggregated bag-level representation to a prediction as:
\begin{equation}\label{eq1}
\hat{Y_i}=f\left(g\left(\left\{x_{i,j}\right\}_{j=1}^{N_i}\right)\right),
\end{equation}
where $g\left(\cdot\right)$ and $f\left(\cdot\right)$ represent the aggregator and the fully connected (FC) layer in the MIL classifier, respectively. 

\subsubsection{Attention-based MIL Methods} In attention-based MIL models, the attention score derived from the pooling operation proposed in \cite{ilse2018attention} is commonly used to measure IIS. Specifically, the attention score, denoted as $\alpha$, is calculated for each instance in the bag, providing a measure of its significance in the overall classification decision. Thus, the attention-based aggregation can be expressed as:
\begin{equation}
g\left(\left\{x_{i,j}\right\}_{j=1}^{N_i}\right) = \sum_{j=1}^{N_i} \alpha_{i,j} \cdot x_{i,j},
\end{equation}
where $\alpha_{i, j}$ represents the attention score assigned to the $j$-th instance in the $i$-th bag. By incorporating these attention scores, the model not only improves its predictive accuracy but also offers insights into which instances most significantly influence the classification outcome.

\subsubsection{Pseudo Bag Augmentation for MIL} Consider the pseudo bag augmented MIL, a regular bag is randomly split into $M$ pseudo bags, and each pseudo bag inherits the label from its parent bag, resulting in an expanded training set $\mathcal{D}^{pse}=\left\{X_i^{pse},Y_i\right\}_{i=1}^{M\times\left|\mathcal{D}\right|}$, where $\left|\mathcal{D}\right|$ is the number of bags. By obtaining ${\hat{Y}}^{pse}$ via Eq. \ref{eq1}, the objective function for pseudo bag augmented MIL is defined:
\begin{equation}\label{eq2}
\mathcal{J}\left(\mathcal{D}^{pse};\theta\right)=\sum_{i=1}^{M\times\left|\mathcal{D}\right|}{{\mathcal{L}\left({\hat{Y}}_{i}^{pse},Y_{i}\right)}},
\end{equation}
where $\theta$ represents the parameter of the MIL classifier, and $\mathcal{L}$ represents the cross-entropy loss function. Nevertheless, the label inherited from the parent bag does not always align with the actual label of the pseudo bag. Thus, the objective function in Eq. \ref{eq2} can be further divided into two parts:
\begin{equation}\label{eq3}
\begin{aligned}
\mathcal{J}\left(\mathcal{D}^{pse};\theta;\varepsilon\right)&=\sum_{i=1}^{M\times\left|\mathcal{D}\right|-\varepsilon}{\mathcal{L}\left(\left.{\hat{Y}}_i^{pse},Y_i\right|{Y_i=Y}_i^{pse}\right)}\\
&+\sum_{i=1}^{\varepsilon}{\mathcal{L}\left(\left.{\hat{Y}}_i^{pse},Y_i\right|{Y_i\neq Y}_i^{pse}\right)},
\end{aligned}
\end{equation}
where $\varepsilon$ is the number of pseudo bags with incorrectly assigned labels. Eq. \ref{eq3} reveals a trade-off between bolstering the diversity of instances and introducing extra noise. 
Existing MIL methods \cite{shao2021weakly,zhang2022dtfd,liu2024pseudo,yang2023protodiv} employ a strategy of randomly splitting bags into pseudo bags, leading to suboptimal outcomes.

\subsection{Shapley Value-based IIS Estimation}
Our observation in Fig. \ref{attn} reveals that $\alpha$ might not accurately reflect the ranking of importance. Thus, we introduce the Shapley value $\phi$ as an alternative method in contrast to the attention score to estimate IIS:
\begin{equation}\label{eq4}
\begin{aligned}
\phi_{i,j}\left(x_{i,j}, X_i\backslash\left\{x_{i,j}\right\}\right)\triangleq &\sum_{S_i\subseteq X_i\backslash\left\{x_{i,j}\right\}}{\frac{\left|S_i\right|!\left(\left|X_i\right|-\left|S_i\right|-1\right)!}{\left|X_i\right|!}}\\
&\times \left[f\left(g\left(S_i\cup\left\{x_{i,j}\right\}\right)\right)-f\left(g\left(S_i\right)\right)\right],
\end{aligned}
\end{equation}
where $x_{i,j}$ is the $j$-th feature in the $i$-th bag to calculate the Shapley value $\phi_{i,j}$, $X_i$ is the full feature set of the $i$-th bag, $S_i \subseteq X_i\backslash\left\{x_{i,j}\right\}$ are all available subsets.

Upon scrutinizing Eq. \ref{eq4}, the computational complexity of the original Shapley value formulation escalates exponentially with the number of instances, which is prohibitively time-intensive in WSI classification as each bag encompasses thousands of instances. To hasten this process, several methodologies have been developed to approximate Shapley values effectively \cite{strumbelj2010efficient,lundberg2017unified,chen2018shapley,ancona2019explaining}. Notably, according to the principle of MIL, it is the positive instances that determine the bag label. Under this premise, we leave the less significant instances in the order of attention scores, and focus on rearranging the importance ranking of instances with high attention scores by their Shapley value-based IIS:
\begin{equation}\label{eq5}
{\rm IIS}\left(x_{i,j}\right)=\phi_{i,j}\left(x_{i,j}, S_i^l\right), x_{i,j}\in S_i^h,
\end{equation}
where $S_i^h$ and $S_i^l=X_i-S_i^h$ denote the instance subset with high and low attention scores, and $X_i$ is the universal set of all instances. For the sake of simplicity, the instance number in $S_i^h$ is set to $\mu M$, and the sampling number for $S_i^l$ is set to $\tau$. Under the assumption that the reasoning time per bag for the model approximates a constant $\gamma$, we quantify the computational complexity of different IIS estimations:

\begin{equation}\label{eq6}
\mathrm{\Omega}\left(\alpha\right)=\sum_{i=1}^{\left|\mathcal{D}\right|}\gamma=\gamma\left|\mathcal{D}\right|,
\end{equation}
\begin{equation}\label{eq7}
\mathrm{\Omega}\left(\phi\left(x, X\backslash\left\{x\right\}\right)\right)=\sum_{i=1}^{\left|\mathcal{D}\right|}\sum_{j=0}^{N_i}{C_{N_i}^j\cdot\gamma}=\gamma\sum_{i=1}^{\left|\mathcal{D}\right|}2^{N_i},
\end{equation}
\begin{equation}\label{eq8}
\mathrm{\Omega}\left(\phi\left(x, S^l\right)\right)=\sum_{i=1}^{\left|S_i^h\right|}\sum_{j=0}^{\tau}\gamma=\gamma\tau\mu M,
\end{equation}
where $\mathrm{\Omega}$ is the asymptotic lower bound of the computational complexity, and $C_{N_i}^j$ is the combination value. Derived from Eq. \ref{eq4}, the calculation of the Shapley value involves exponential computational complexity, while that of the attention score exhibits linear complexity. By employing our approximation technique, we are capable of transforming the Shapley-based IIS computation into a linear complexity as depicted in Eq. \ref{eq8}, while ensuring its ranking accuracy remains intact within the realm of multiple-instance learning.
\begin{table*}[]
\caption{Bag-level performance results for 5 repeatability experiments on CAMELYON-16, BRACS, TCGA-LUNG, and TCGA-BRCA test sets. The subscripts are the standard deviation of each metric. The best evaluation results are in bold.}
\label{table1}
\centering
\renewcommand\arraystretch{1.2}
\resizebox{\textwidth}{!}{
\begin{tabular}{ccccccccccccc}
\toprule
\multirow{2}{*}{Method} & \multicolumn{3}{c}{CAMELYON-16} & \multicolumn{3}{c}{BRACS} & \multicolumn{3}{c}{TCGA-LUNG} & \multicolumn{3}{c}{TCGA-BRCA} \\ \cmidrule(r){2-4} \cmidrule(r){5-7} \cmidrule(r){8-10} \cmidrule(r){11-13}
& ACC(\%) & AUC(\%) & F1(\%) & ACC(\%) & AUC(\%) & F1(\%) & ACC(\%) & AUC(\%) & F1(\%) & ACC(\%) & AUC(\%) & F1(\%) \\
\midrule
MeanMIL & 70.9\textsubscript{1.8} & 58.7\textsubscript{1.9} & 62.3\textsubscript{3.3} & 52.4\textsubscript{2.6} & 69.2\textsubscript{1.6} & 40.6\textsubscript{2.5} & 82.0\textsubscript{0.9} & 88.9\textsubscript{2.0} & 82.0\textsubscript{1.0} & 84.0\textsubscript{2.2} & 84.7\textsubscript{5.1} & 72.1\textsubscript{5.7} \\
MaxMIL & 83.7\textsubscript{1.8} & 86.7\textsubscript{2.6} & 83.3\textsubscript{5.1} & 55.9\textsubscript{2.8} & 75.9\textsubscript{1.6} & 50.3\textsubscript{4.0} & 88.7\textsubscript{1.0} & 94.4\textsubscript{1.2} & 88.7\textsubscript{1.0} & 84.5\textsubscript{1.8} & 84.6\textsubscript{4.7} & 70.4\textsubscript{6.3} \\
ABMIL \cite{ilse2018attention} & 82.5\textsubscript{1.9} & 83.8\textsubscript{2.1} & 80.6\textsubscript{1.7} & 58.4\textsubscript{0.9} & 76.1\textsubscript{0.6} & 54.7\textsubscript{2.3} & 87.6\textsubscript{0.7} & 93.1\textsubscript{1.8} & 87.6\textsubscript{0.7} & 84.9\textsubscript{2.3} & 82.7\textsubscript{4.8} & 75.8\textsubscript{3.6} \\
DSMIL \cite{li2021dual} & 77.2\textsubscript{1.7} & 77.2\textsubscript{2.1} & 74.4\textsubscript{2.6} & 53.1\textsubscript{2.2} & 70.8\textsubscript{3.3} & 46.1\textsubscript{3.7} & 86.2\textsubscript{1.4} & 93.6\textsubscript{1.0} & 86.2\textsubscript{1.4} & 84.4\textsubscript{1.1} & 83.2\textsubscript{5.2} & 73.4\textsubscript{3.9} \\
CLAM \cite{lu2021data} & 82.5\textsubscript{3.2} & 81.6\textsubscript{2.4} & 80.1\textsubscript{3.5} & 53.8\textsubscript{3.5} & 73.3\textsubscript{1.7} & 51.5\textsubscript{3.3} & 88.2\textsubscript{1.4} & 94.2\textsubscript{1.2} & 88.2\textsubscript{1.4} & 85.9\textsubscript{2.8} & 85.8\textsubscript{4.2} & 77.0\textsubscript{5.3} \\
TransMIL \cite{shao2021transmil} & 85.0\textsubscript{1.4} & 89.1\textsubscript{0.7} & 83.3\textsubscript{1.3} & 57.0\textsubscript{2.4} & 75.5\textsubscript{1.0} & 49.2\textsubscript{5.2} & 87.9\textsubscript{0.9} & 94.8\textsubscript{0.8} & 87.9\textsubscript{0.9} & 83.4\textsubscript{2.1} & 83.4\textsubscript{4.3} & 68.1\textsubscript{6.5} \\
DTFD \cite{zhang2022dtfd} & 85.3\textsubscript{1.6} & 85.4\textsubscript{3.2} & 84.9\textsubscript{1.7} & 57.2\textsubscript{2.7} & 76.6\textsubscript{2.0} & 56.2\textsubscript{3.8} & 88.8\textsubscript{0.6} & 94.6\textsubscript{0.8} & 88.8\textsubscript{0.6} & 85.6\textsubscript{2.3} & 84.6\textsubscript{4.1} & 76.4\textsubscript{4.1} \\
PseMix \cite{liu2024pseudo} & 85.6\textsubscript{3.3} & 86.1\textsubscript{4.4} & 78.3\textsubscript{5.5} & 57.9\textsubscript{3.5} & 77.3\textsubscript{1.9} & 53.2\textsubscript{5.6} & 89.8\textsubscript{0.6} & 95.2\textsubscript{1.3} & 89.5\textsubscript{0.9} & 84.9\textsubscript{1.6} & 84.4\textsubscript{3.6} & 61.5\textsubscript{2.7} \\
Additive MIL \cite{javed2022additive} & 65.3\textsubscript{6.1} & 56.4\textsubscript{3.4} & 59.6\textsubscript{4.8} & 54.3\textsubscript{2.6} & 74.2\textsubscript{0.5} & 47.4\textsubscript{1.9} & 80.9\textsubscript{3.6} & 88.3\textsubscript{1.4} & 80.8\textsubscript{3.8} & 83.5\textsubscript{1.9} & 81.8\textsubscript{4.4} & 72.4\textsubscript{4.0} \\
PMIL & \textbf{87.4}\textsubscript{1.1} & \textbf{90.1}\textsubscript{1.6} & \textbf{86.3}\textsubscript{1.1} & \textbf{67.1}\textsubscript{3.3} & \textbf{82.8}\textsubscript{1.8} & \textbf{66.4}\textsubscript{3.1} & \textbf{91.3}\textsubscript{1.4} & \textbf{96.5}\textsubscript{0.9} & \textbf{91.3}\textsubscript{1.4} & \textbf{86.9}\textsubscript{1.5} & \textbf{86.3}\textsubscript{4.2} & \textbf{79.0}\textsubscript{3.0} \\
\bottomrule
\end{tabular}}
\end{table*}
\subsection{IIS-based Pseudo Bag Augmentation}
As illustrated in Fig. \ref{overview}(b), instances within each bag are rearranged according to the ranking of IIS, denoted as $X_i^\prime=\left\{\left.x_{i,j}^\prime\right|\ {\rm IIS}\left(x_{i,1}^\prime\right)\geq {\rm IIS}\left(x_{i,2}^\prime\right)\geq\cdots\geq {\rm IIS}\left(x_{i,N_i}^\prime\right)\right\}$. And these instances are evenly interleaved into $M$ pseudo bags by using the modulo function $mod$ to constrain ${x_{i,j}^\prime}$ to satisfy $j\equiv k\ (\text{mod}\ M)$, resulting in that each pseudo bag $X_{i,k}^{pse}$ denotes as a sample from $X_i^\prime$. By fixing the parameter $\theta$ of MIL models, the optimization of $\varepsilon$ can be approximated as:
\begin{equation}\label{eq10}
\begin{aligned}
\varepsilon^{\ast} &= \mathop{\min}\sum \left({\hat{Y}}_{i} \neq Y_{i}^{pse}\right) = \mathop{\max}\sum \left({\hat{Y}}_{i} = Y_{i}^{pse}\right)\\
&\iff\mathop{\min}_{X^\prime}{D_{KL}\left(P_{X^{pse} \sim \Gamma\left(X^\prime\right)}\left[\left.Y^{pse}\right|X^{pse};\theta \right]\right.} \\
&\qquad \qquad \qquad \quad \left.\parallel P\left[\left.Y\right|X;\theta\right]\right),
\end{aligned}
\end{equation}
where $\Gamma\left(X^\prime\right)$ is the instance importance distribution of $X^\prime$ with estimated IIS and $D_{KL}$ is Kullback-Leibler divergence function. Thus, the optimization of $\varepsilon$ is translated to that of $\Gamma\left(X^\prime\right)$, where the IIS estimation plays a decisive role.

Furthermore, it is important to consider progressive strategies concerning the quantity and initialization of pseudo bags. Splitting a regular bag into a large number of pseudo bags can introduce excessive noise, which may lead to training instability, particularly when regular bags contain only a limited number of positive instances. To address this issue, we progressively increase the number of pseudo bags once the MIL model converges during training:
\begin{equation}\label{eq11}
\begin{aligned}
M_t=\min\left \{M_{t-1}+\Delta M,M_{max}\right \},&\\
s.t.\left \{g_{t-1},f_{t-1}\right \}\to &\left \{g_{t-1}^*,f_{t-1}^*\right \},
\end{aligned}
\end{equation}
where $t$ signifies the convergence iteration, $\Delta M$ denotes the increment in the number of pseudo bags, and $M_0$ and $M_{\text{max}}$ represent the initial and maximum numbers of pseudo bags, respectively. In addition, the initial assignment of pseudo bags significantly influences subsequent training, especially when dealing with challenging datasets. To address this issue, we gradually leverage the well-trained MIL model from the previous round to enhance the initial pseudo bag augmentation by calculating instance importance scores.

\subsection{Progressive Pseudo Bag Augmented MIL}
Under the guidance of IIS estimated by Shapley values instead of attention scores used in the existing architectures, we propose a progressive pseudo bag augmented MIL framework termed PMIL. To alleviate the mislabeling issue in Eq. \ref{eq3}, we incorporate the expectation-maximization (EM) algorithm \cite{dempster1977maximum} to obtain optimal pseudo bag label assignment. Specifically, the parameter $\theta$ of the MIL model can be learned via Eq. \ref{eq2} as the M-step, and the minimization problem for $\varepsilon$ is translated into an assignment optimization for pseudo bags via Eq. \ref{eq10} as the E-step. As illustrated in Fig. \ref{overview}(b), this iterative optimization is implemented within $n$ training rounds.

\begin{table*}[]
\caption{Instance-level performance results for 5 repeatability experiments on CAMELYON-16 test set. The subscripts are the standard deviation of each metric. The best evaluation results are in bold.}
\label{instance_infer}
\renewcommand\arraystretch{1.2}
\centering
\resizebox{0.72\textwidth}{!}{
\begin{tabular}{ccccccc}
\toprule
\multirow{2}{*}{Method} &
  \multicolumn{5}{c}{Instance-level Evaluation Metrics} &
  \multirow{2}{*}{Average(\%)} \\
 &
  ACC(\%) &
  AUC(\%) &
  F1(\%) &
  Precision(\%) &
  Recall(\%) &
   \\
 \midrule
MeanMIL &
  83.65\textsubscript{5.29} &
  89.83\textsubscript{1.37} &
  61.87\textsubscript{7.34} &
  49.68\textsubscript{9.42} &
  84.18\textsubscript{2.40} &
  73.84 \\
MaxMIL &
  89.63\textsubscript{0.62} &
  \textbf{95.49}\textsubscript{0.30} &
  47.78\textsubscript{4.89} &
  99.78\textsubscript{0.07} &
  31.55\textsubscript{4.11} &
  72.85 \\
ABMIL \cite{ilse2018attention} &
  56.23\textsubscript{1.70} &
  73.64\textsubscript{1.58} &
  39.69\textsubscript{0.78} &
  25.08\textsubscript{0.66} &
  \textbf{95.13}\textsubscript{0.64} &
  57.96 \\
DSMIL \cite{li2021dual} &
  87.64\textsubscript{0.49} &
  93.77\textsubscript{0.43} &
  30.84\textsubscript{4.73} &
  \textbf{99.91}\textsubscript{0.04} &
  18.33\textsubscript{3.25} &
  66.10 \\
CLAM \cite{li2021dual} &
  64.54\textsubscript{15.68} &
  78.41\textsubscript{11.43} &
  44.26\textsubscript{14.47} &
  32.12\textsubscript{14.95} &
  80.62\textsubscript{8.10} &
  59.99 \\
TransMIL \cite{shao2021transmil} &
  40.47\textsubscript{3.58} &
  68.75\textsubscript{2.17} &
  32.57\textsubscript{1.16} &
  19.67\textsubscript{0.87} &
  94.83\textsubscript{0.60} &
  51.26 \\
DTFD \cite{zhang2022dtfd} &
  63.37\textsubscript{21.41} &
  83.14\textsubscript{7.17} &
  47.42\textsubscript{18.10} &
  37.55\textsubscript{22.07} &
  83.35\textsubscript{5.52} &
  62.97 \\
PMIL &
  \textbf{93.75}\textsubscript{0.34} &
  92.82\textsubscript{0.97} &
  \textbf{74.85}\textsubscript{2.53} &
  95.39\textsubscript{3.69} &
  61.98\textsubscript{5.38} &
  \textbf{83.76}\\
\bottomrule
\end{tabular}}
\end{table*}
\section{Experiments}
\subsection{Datasets and Evaluation Metrics}
Our experimental setup employs four publicly available datasets to assess the performance of our proposed method.

\textbf{CAMELYON-16} focuses on detecting lymph node metastasis in early-stage breast cancer. It comprises 399 WSIs, with 270 allocated for training and 129 for testing. The official training set follows a 5-fold cross-validation protocol to generate training and validation sets. Furthermore, a total of 337,124 negative instances and 60,077 positive instances are assigned instance labels based on the annotations in the test set for subsequent evaluation. 

\textbf{BRACS} \cite{brancati2022bracs} is curated for breast cancer subtyping and contains 547 WSIs. The classification task involves benign tumors, atypical tumors (AT), and malignant tumors (MT). We adhere to the official dataset split, with 395 for training, 65 for validating, and 87 for testing. We conduct five separate experiments with different random seeds.

\textbf{TCGA-LUNG} comprises 1034 WSIs, encompassing 528 lung adenocarcinoma (LUAD) and 506 lung squamous cell carcinoma (LUSC) cases. We adopt a 5-fold cross-validation protocol for both training and testing.

\textbf{TCGA-BRCA} comprises 985 WSIs, encompassing 797 invasive ductal carcinoma (IDL) and 198 invasive lobular carcinoma (ILC) cases. We adopt a 5-fold cross-validation protocol for both training and testing.
\begin{figure*}[]
  \centering
  \setlength{\abovecaptionskip}{0pt}%
  \setlength{\belowcaptionskip}{0pt}%
  \includegraphics[width=0.95\textwidth]{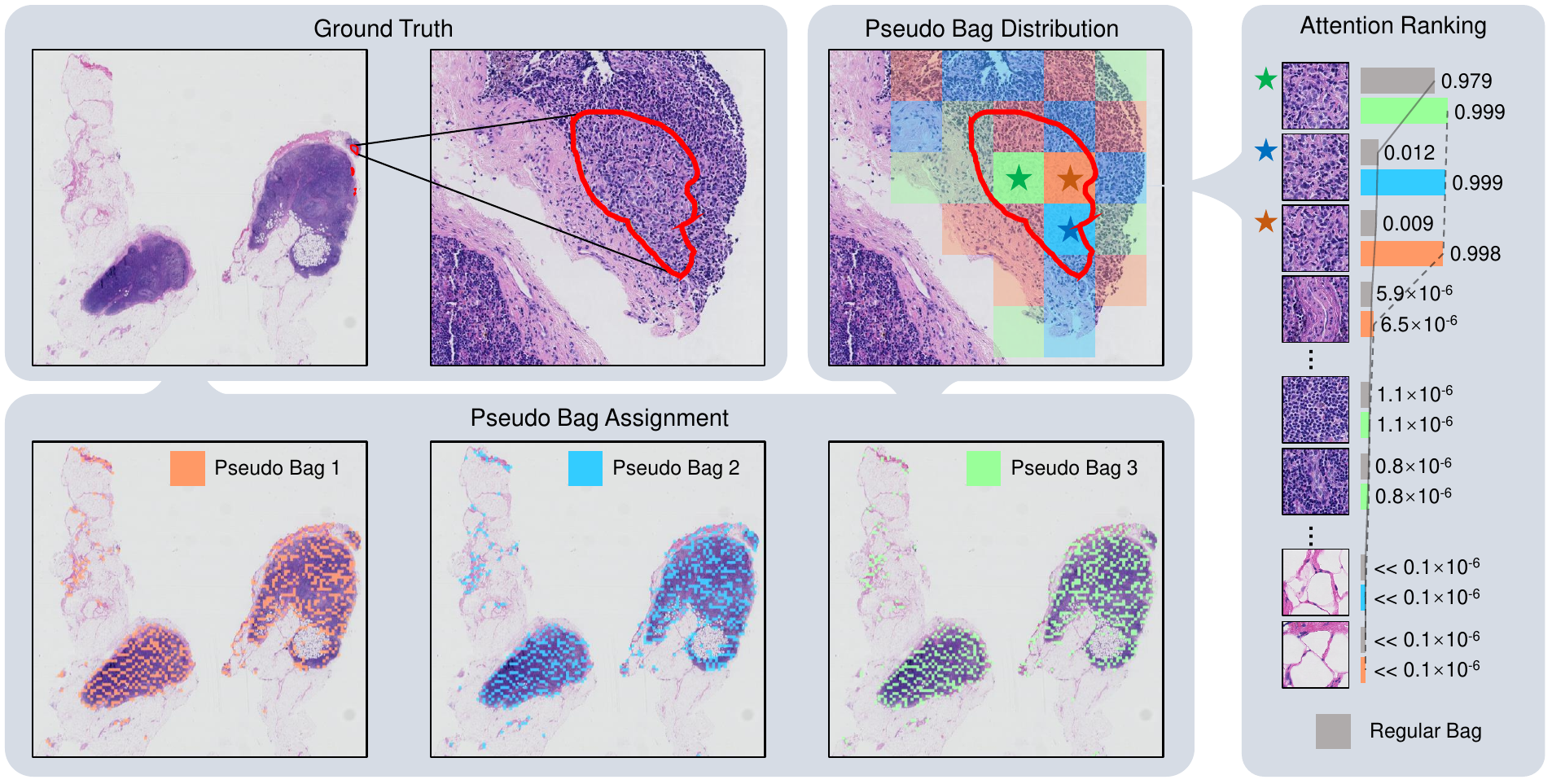}
  \caption{Visualization of pseudo bag assignment using PMIL. The red annotations represent cancer regions. Our method can locate only three positive instances even in the micro metastasis case based on the ranking of Shapley values, and split them into pseudo bags evenly. The attention ranking reveals that more positive instances are noticed during training by accurate pseudo bag augmentation.}\label{pseudobag}
\end{figure*}

For bag-level evaluation, we present multi-class evaluation metrics: slide-level accuracy (ACC), one-versus-rest area under the curve (AUC), and macro F1 score.

For instance-level evaluation, we present binary evaluation metrics: ACC, AUC, F1 score, precision, and recall.
\subsection{Implementation Details}
In the preprocessing stage, we utilize OTSU's thresholding method to detect and localize tissue regions for patch generation. We create non-overlapping patches measuring 256$\times$256 pixels at magnifications of 20$\times$ for CAMELYON-16, TCGA-LUNG, and TCGA-BRCA, and 5$\times$ for BRACS. This process results in an average of approximately 7156, 11951, 2392, and 714 patches per bag for these datasets, respectively.

All experiments were performed on a workstation equipped with NVIDIA RTX 3090 GPUs. We employed ResNet50 \cite{he2016deep} as the encoder and ABMIL \cite{ilse2018attention} as the primary MIL model. The Adam optimizer, with a weight decay of 1e-5, was selected. We also implemented an early stopping strategy, setting the patience parameter to 20 epochs. The initial learning rate was established at 3e-4 and subsequently reduced to 1e-4 for fine-tuning purposes. For the CAMELYON-16 dataset, we limited the maximum number of pseudo bags to 8; for BRACS, TCGA-LUNG, and TCGA-BRCA, the limits were 10, 14, and 6, respectively. The increment in the number of pseudo bags is set to 4. In terms of Shapley value computation acceleration, we set the parameter $\mu$ to 10 and $\tau$ to 3. The total EM training round $n$ is set to 10.
\subsection{Evaluation and Comparison}
We present the experimental results of our proposed PMIL framework built on the ABMIL \cite{ilse2018attention} backbone for CAMELYON-16, BRACS, TCGA-LUNG, and TCGA-BRCA datasets, comparing them with the following methods: (1) Conventional instance-level MIL, including the Mean-Pooling MIL and Max-Pooling MIL. (2) The vanilla attention-based MIL, ABMIL \cite{ilse2018attention}. (3) Two variants of ABMIL, including non-local attention pooling, DSMIL \cite{li2021dual}, and single-attention-branch, CLAM-SB \cite{lu2021data}. (4) Transformer-based MIL, TransMIL \cite{shao2021transmil}. (5) Pseudo bag augmented MIL, DTFD \cite{zhang2022dtfd} and PseMix \cite{liu2024pseudo}. (6) Shapley value enabled MIL, Additive MIL.

\begin{figure*}[]
  \centering
  \setlength{\abovecaptionskip}{0pt}%
  \setlength{\belowcaptionskip}{0pt}%
  \includegraphics[width=\textwidth]{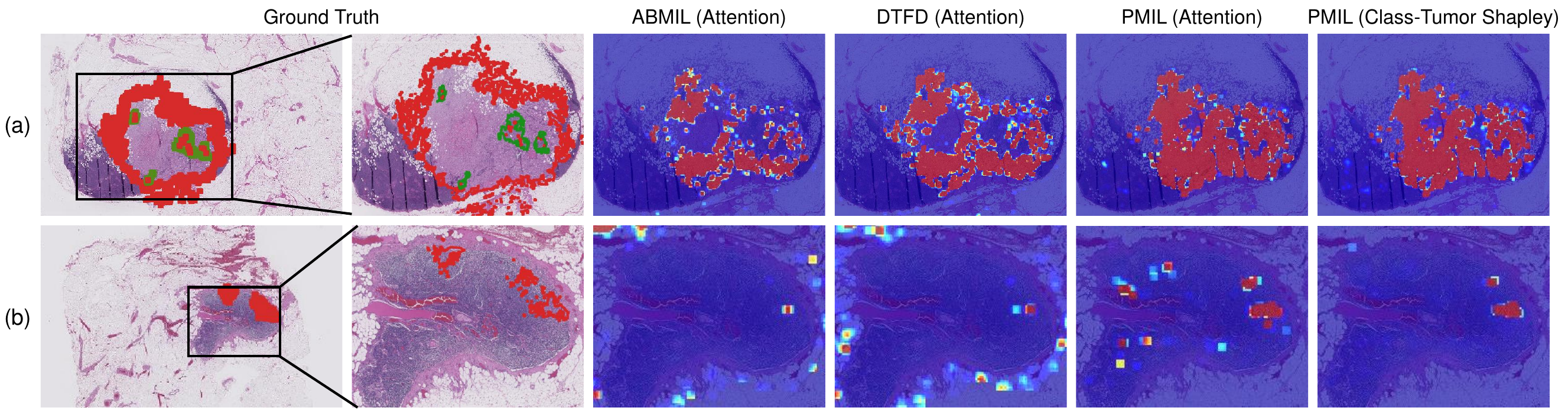}
    \caption{Heatmaps of two slide sub-fields using different models. (a) and (b) are macro and micro metastasis cases from CAMELYON-16, where red and green annotations are cancer and noncancer regions in the column of 'Ground Truth'.} \label{visualization}
\end{figure*}

\begin{figure}[]
  \centering
  \setlength{\abovecaptionskip}{0pt}%
  \setlength{\belowcaptionskip}{0pt}%
  \includegraphics[width=0.8\linewidth]{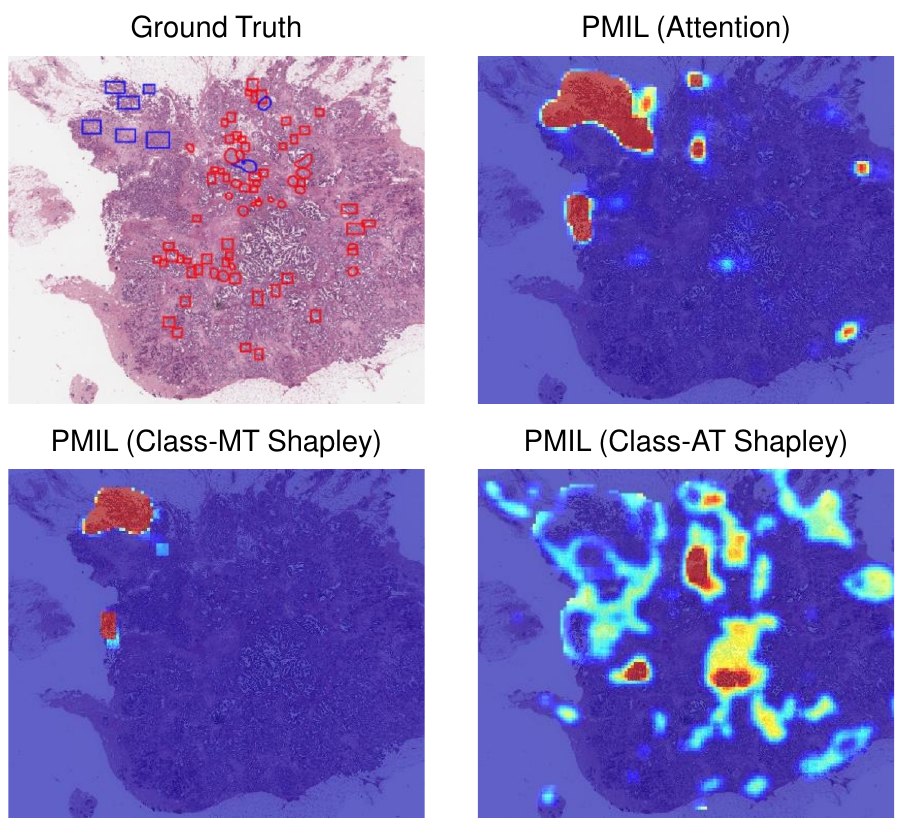}
    \caption{Heatmaps of one malignant tumor (MT) case from BRACS with various IIS estimations. The blue and red annotations are the malignant tumor and atypical tumor (AT) regions in the 'Ground Truth'.} \label{bracs_vis}
\end{figure}
In the bag-level evaluation, as shown in Table \ref{table1}, our proposed PMIL framework demonstrates remarkable performance, achieving AUC scores of 90.1\% for CAMELYON-16, 82.8\% for BRACS, 96.5\% for TCGA-LUNG, and 86.3\% for TCGA-BRCA. These scores consistently exceed those of all other methods included in the comparison. Notably, on the complex BRACS dataset, our approach exhibits significant superiority. The generation of progressively refined pseudo bags contributes to enhanced training diversity and a reduction in the number of instances per bag. This strategy effectively improves the proficiency of the model in learning from positive instances.

In the instance-level evaluation, as illustrated in Table \ref{instance_infer}, certain methods, such as MaxMIL and DSMIL, exhibited high precision scores but low recall scores, indicating a cautious tendency towards predicting positive instances. Conversely, methods like MeanMIL, ABMIL, CLAM, TransMIL, and DTFD displayed the opposite trend, often predicting a larger number of instances with positive labels, albeit less precisely. In contrast, our method provided significantly more precise predictions for positive instances and outperformed other methods in terms of both ACC and F1 score.
\subsection{Visualization and Interpretation}
To assess the effectiveness of our progressive pseudo bag augmentation in shifting the network's focus toward more positive instances, we analyzed the attention distribution of our method. Fig. \ref{attn}(d) shows that PMIL achieves a more evenly spread attention distribution compared to ABMIL, CLAM, and DTFD. Additionally, the total of the top 10 attention scores is reduced to 0.36 in normal cases and to 0.53 in tumor cases. A specific instance of pseudo bag assignment in PMIL is depicted in Fig. \ref{pseudobag}. In the case of micro-metastasis, PMIL successfully identifies three crucial patches. The random partitioning approach only has a chance with a rate of 2/9 to accurately allocate positive instances across three different pseudo bags, which could otherwise contribute noise to the training. In contrast, our method confidently places these patches into different pseudo bags, significantly increasing the diversity of positive instances. 

To emphasize the limitations of the attention score-based IIS, we conducted a comparative analysis, as illustrated in Fig. \ref{visualization}(a) and (b). In cases of macro metastasis, ABMIL, DTFD, and our model show effective performance. However, in micro metastasis scenarios, the attention score-based IIS suggests that ABMIL, DTFD, and our model erroneously focus on some noncancerous areas, which eludes logical interpretation. Conversely, using Shapley value-based IIS, our model precisely excludes noncancerous regions and accurately pinpoints cancerous areas.

\begin{figure*}[]
  \centering
  \setlength{\abovecaptionskip}{0pt}%
  \setlength{\belowcaptionskip}{0pt}%
  \includegraphics[width=\textwidth]{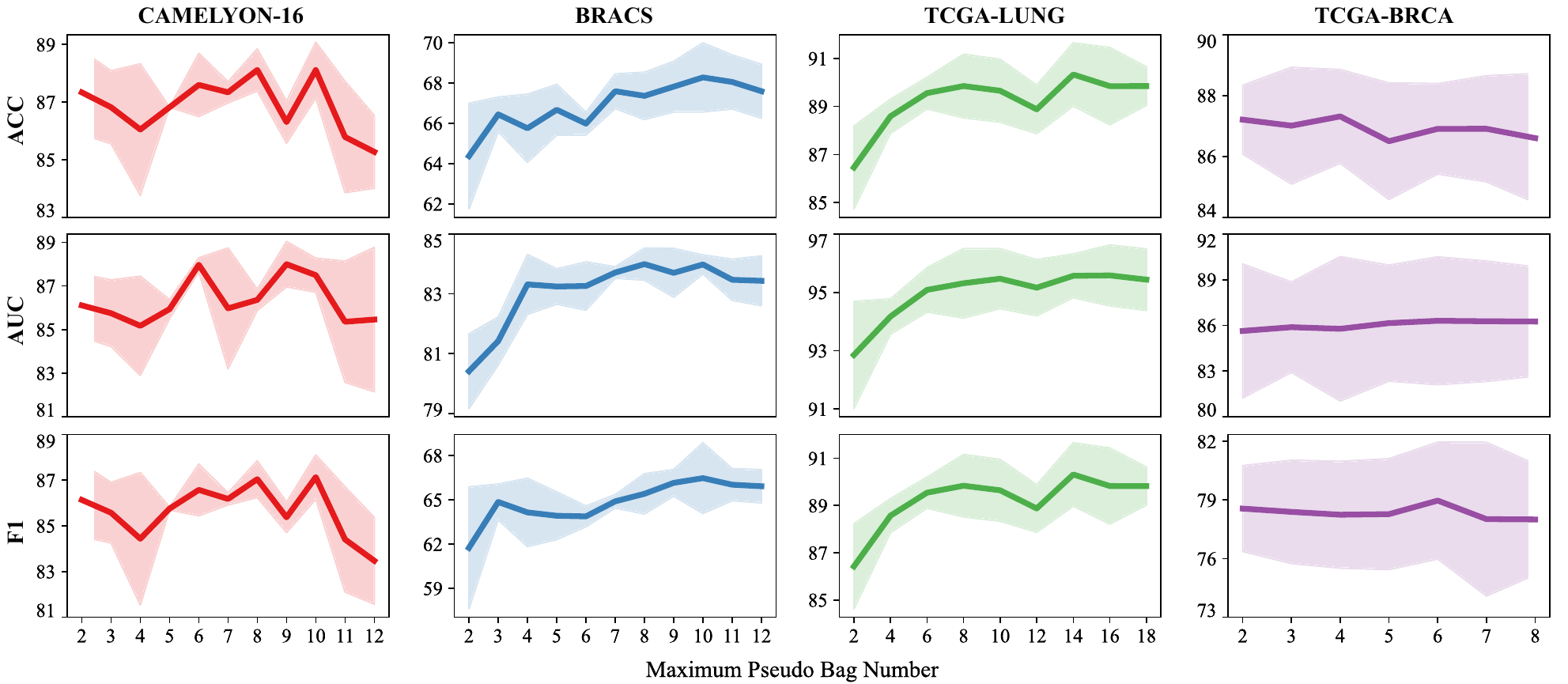}
  \caption{Ablation results of different maximum pseudo bag numbers $M_{max}$ on CAMELYON-16, BRACS, TCGA-LUNG, and TCGA-BRCA datasets.} \label{maxnum}
\end{figure*}

\begin{table*}[]
\caption{Performance results of pseudo bag augmented MIL using different IIS measurements on CAMELYON-16, BRACS, TCGA-LUNG, and TCGA-BRCA test sets. The subscripts are the standard deviation. The best evaluation results are in bold.}
\label{table2}
\centering
\renewcommand\arraystretch{1.2}
\resizebox{\textwidth}{!}{
\begin{tabular}{ccccccccccccc}
\toprule
\multirow{2}{*}{Metrics} & \multicolumn{3}{c}{CAMELYON-16} & \multicolumn{3}{c}{BRACS} & \multicolumn{3}{c}{TCGA-LUNG} & \multicolumn{3}{c}{TCGA-BRCA}\\ \cmidrule(r){2-4} \cmidrule(r){5-7} \cmidrule(r){8-10} \cmidrule(r){11-13}
 & ACC & AUC & F1 & ACC & AUC & F1 & ACC & AUC & F1 & ACC & AUC & F1 \\
\midrule
Random & $87.1\textsubscript{0.9}$ & $86.2\textsubscript{2.5}$ & $85.5\textsubscript{1.2}$ & $61.8\textsubscript{2.4}$ & $80.4\textsubscript{0.5}$ & $58.5\textsubscript{5.9}$ & $89.7\textsubscript{1.9}$ & $95.8\textsubscript{1.0}$ & $89.6\textsubscript{1.9}$ & $86.8\textsubscript{1.9}$ & $85.6\textsubscript{3.7}$ & $77.9\textsubscript{4.0}$\\
Attention Score & $\textbf{87.4}\textsubscript{2.8}$ & $89.8\textsubscript{0.7}$ & $86.2\textsubscript{2.7}$ & $\textbf{68.3}\textsubscript{1.7}$ & $\textbf{84.0}\textsubscript{0.3}$ & $\textbf{66.5}\textsubscript{2.4}$ & $90.3\textsubscript{1.3}$ & $95.6\textsubscript{0.8}$ & $90.3\textsubscript{1.3}$ & $86.5\textsubscript{1.8}$ & $85.5\textsubscript{3.9}$ & $77.2\textsubscript{2.7}$\\
Shapley Value & $\textbf{87.4}\textsubscript{1.1}$ & $\textbf{90.1}\textsubscript{1.6}$ & $\textbf{86.3}\textsubscript{1.1}$ & $67.1\textsubscript{3.3}$ & $82.8\textsubscript{1.8}$ & $66.4\textsubscript{3.1}$ & $\textbf{91.3}\textsubscript{1.2}$ & $\textbf{96.5}\textsubscript{0.8}$ & $\textbf{91.3}\textsubscript{1.2}$ & $\textbf{86.9}\textsubscript{1.5}$ & $\textbf{86.3}\textsubscript{4.2}$ & $\textbf{79.0}\textsubscript{3.0}$\\
\bottomrule
\end{tabular}}
\end{table*}
Distinct from attention scores, the computation of Shapley values encompasses the entire MIL classifier, incorporating diverse category information, thereby enabling interpretations on a class-wise basis. As shown in Fig. \ref{bracs_vis}, the IIS estimated by attention scores and class-MT Shapley values predominantly focuses on malignant tumor regions. Meanwhile, the heatmaps generated using class-AT Shapley values predominantly emphasize atypical tumor regions, aligning with the slide-level labels. Although the heatmaps might not be entirely accurate for the BRACS dataset, this finding highlights the robust interpretability of Shapley value-based IIS in multiple classification tasks.

In summary, the visualization results indicate that while attention score-based IIS often produces a noisy ranking of instance importance and is limited to a single target category, Shapley value-based IIS ensures a more accurate ranking of instance importance and enables class-wise interpretations, leveraging the full capacity of the MIL classifier.

\subsection{Ablation study}
\subsubsection{IIS Measure Estimation Metrics}
In this ablation study, we evaluated both the attention score and the Shapley value as methods for estimating IIS for subsequent training, using random splitting as the baseline for pseudo bag augmentation. According to the results presented in Table \ref{table2}, Shapley value-based IIS estimation demonstrates superior performance on the CAMELYON-16, TCGA-LUNG, and TCGA-BRCA datasets. Conversely, on the BRACS dataset, the attention score-based estimation yielded better results. This variation in effectiveness is likely due to the direct acquisition of attention scores via pooling operations. In contrast, calculating the Shapley value requires an additional fully connected layer, which may be less robust when the overall performance of the MIL classifier is not particularly high. Therefore, the Shapley value estimation is more advantageous with datasets that pose fewer learning challenges.

\begin{table}[]
\centering
\caption{AUC results of the proposed method using different hyper-parameters on CAMELYON-16, BRACS, TCGA-LUNG and TCGA-BRCA test sets. The subscripts are the standard deviation. The best evaluation results are in bold.}
\label{hyperparam}
\renewcommand\arraystretch{1.2}
\resizebox{\linewidth}{!}{
\begin{tabular}{cccccc}
\toprule
$\mu$ & $\tau$ & CAMELYON-16 & BRACS & TCGA-LUNG & TCGA-BRCA \\ \midrule
5 & \multirow{4}{*}{3} & 90.51\textsubscript{0.57} & 82.38\textsubscript{1.84} & 95.48\textsubscript{0.14} & 85.31\textsubscript{4.99} \\
10 &  & 90.10\textsubscript{1.61} & 82.82\textsubscript{1.84} & 95.74\textsubscript{0.27} & 86.91\textsubscript{4.35} \\
15 &  & 90.00\textsubscript{0.93} & 82.61\textsubscript{1.50} & 95.71\textsubscript{0.53} & 86.49\textsubscript{4.64} \\
20 &  & 90.40\textsubscript{1.16} & 82.96\textsubscript{1.02} & 95.69\textsubscript{0.48} & 86.38\textsubscript{3.30} \\
\multirow{5}{*}{10} & 1 & 90.33\textsubscript{1.26} & 82.29\textsubscript{1.24} & 95.69\textsubscript{0.71} & 85.35\textsubscript{3.40} \\
 & 2 & 90.94\textsubscript{0.97} & 82.64\textsubscript{0.76} & 95.57\textsubscript{0.95} & 87.07\textsubscript{4.78} \\
 & 3 & 90.10\textsubscript{1.61} & 82.82\textsubscript{1.84} & 95.74\textsubscript{0.27} & 86.91\textsubscript{4.35} \\
 & 4 & 89.52\textsubscript{0.61} & 82.65\textsubscript{1.77} & 96.26\textsubscript{0.88} & 85.92\textsubscript{3.78} \\
 & 5 & 91.04\textsubscript{1.55} & 82.89\textsubscript{1.13} & 96.08\textsubscript{0.32} & 85.67\textsubscript{6.43} \\
\bottomrule
\end{tabular}
}
\end{table}

\begin{table*}[]
\caption{Evaluation of pseudo bag augmentation using different progressive strategies on CAMELYON-16, BRACS, and TCGA-LUNG test sets. The subscripts are the standard deviation. The best evaluation results are in bold.}
\label{table3}
\centering
\renewcommand\arraystretch{1.2}
\resizebox{\linewidth}{!}{
\begin{tabular}{cccccccccccccc}
\toprule
\multicolumn{2}{c}{Pseudo Bag Strategy} & \multicolumn{3}{c}{CAMELYON-16} & \multicolumn{3}{c}{BRACS} & \multicolumn{3}{c}{TCGA-LUNG} & \multicolumn{3}{c}{TCGA-BRCA}\\ \cmidrule(r){1-2} \cmidrule(r){3-5} \cmidrule(r){6-8} \cmidrule(r){9-11} \cmidrule(r){12-14}
Number & Initialization & ACC & AUC & F1 & ACC & AUC & F1 & ACC & AUC & F1 & ACC & AUC & F1 \\
\midrule
Constant & Constant & 80.82\textsubscript{1.0} & 77.51\textsubscript{1.7} & 76.93\textsubscript{1.6} & 62.65\textsubscript{1.3} & 82.78\textsubscript{0.9} & 60.89\textsubscript{1.8} & 90.57\textsubscript{1.4} & 95.92\textsubscript{0.3} & 90.55\textsubscript{1.4} & 85.48\textsubscript{2.0} & 85.80\textsubscript{3.4} & 73.90\textsubscript{6.0} \\
Progressive & Constant & 84.89\textsubscript{2.8} & 85.48\textsubscript{1.6} & 82.41\textsubscript{3.8} & 64.66\textsubscript{2.1} & 82.25\textsubscript{0.8} & 62.12\textsubscript{2.6} & 89.74\textsubscript{2.3} & 95.55\textsubscript{1.6} & 89.72\textsubscript{2.3} & 86.50\textsubscript{1.7} & 85.89\textsubscript{3.7} & 77.88\textsubscript{3.6} \\
Constant & Progressive & 85.08\textsubscript{4.2} & 86.41\textsubscript{4.1} & 83.33\textsubscript{5.4} & 70.69\textsubscript{0.6} & 84.49\textsubscript{0.7} & 68.71\textsubscript{0.8} & 90.57\textsubscript{0.3} & 96.06\textsubscript{0.6} & 90.56\textsubscript{0.3} & 86.29\textsubscript{2.1} & 85.83\textsubscript{3.9} & 77.67\textsubscript{3.3} \\
Progressive & Progressive & \textbf{88.18}\textsubscript{1.2} & \textbf{88.10}\textsubscript{1.0} & \textbf{86.99}\textsubscript{1.5} & \textbf{71.26}\textsubscript{1.2} & \textbf{84.88}\textsubscript{0.2} & \textbf{69.86}\textsubscript{1.4} & \textbf{91.30}\textsubscript{1.7} & \textbf{96.09}\textsubscript{1.3} & \textbf{91.28}\textsubscript{1.8} & \textbf{86.80}\textsubscript{1.7} & \textbf{86.03}\textsubscript{4.3} & \textbf{78.65}\textsubscript{3.0}\\
\bottomrule
\end{tabular}
}
\end{table*}
\subsubsection{Sensitivity to Hyper-parameters}
The optimal number of pseudo bags, $M_{max}$, varies across datasets due to differences in magnification levels of the patches and the sizes of tumor regions. As depicted in Fig. \ref{maxnum}, on the CAMELYON-16 dataset, our model exhibits peak ACC and AUC performance with an $M_{max}$ of approximately 8. Similarly, the ideal $M_{max}$ or the BRACS, TCGA-LUNG, and TCGA-BRCA datasets are found to be 10, 14, and 6, respectively. The smaller $M_{max}$ on CAMELYON-16 can be attributed to the prevalence of micro metastasis slides containing few positive instances, even at a 20$\times$ magnification. In such scenarios, pseudo bag augmentation must balance between introducing additional noise to the training set and enhancing training diversity. Conversely, the larger cancer (subtype) regions in the BRACS, TCGA-LUNG, and TCGA-BRCA datasets allow for division into more pseudo bags without compromising stability.

We also present the repeatability results of various hyper-parameters $\mu$ and $\tau$ utilized in Shapley value acceleration, as depicted in Table \ref{hyperparam}. It indicates that the performance of our proposed approach is not significantly affected by changes in $\mu$ and $\tau$, as the performance fluctuations fall within an acceptable range.
\subsubsection{Progressive Pseudo Bag Augmentation}
To ascertain the efficacy of progressively increasing the pseudo bag count and refining the initial pseudo bag assignment, we carried out a series of experiments. For this, we set the pseudo bag increment $\Delta M$ to 4, with the training rounds, $n$ as 5 for the CAMELYON-16, TCGA-LUNG, and TCGA-BRCA datasets, and extended to 10 for the more intricate BRACS dataset. The results, as indicated in Table \ref{table3}, show that models incorporating both progressive tactics achieve the highest levels of performance. The CAMELYON-16 dataset is particularly sensitive to the number of pseudo bags, necessitating precise calibration to prevent the introduction of undue noise. Conversely, the BRACS dataset's sensitivity lies in the initial setup of pseudo bags, owing to the challenge of distinguishing between subtypes. A more sophisticated initial setup significantly aids the model in accurately recognizing positive instances, leading to enhanced performance.

From these ablation studies, we summarize several key insights as follows:

\textbf{Selection of IIS Estimation Metrics.} The selection of IIS estimation methods depends on the characteristics of the dataset. While attention score-based IIS is commonly employed, its ranking accuracy can sometimes be compromised. In contrast, Shapley value-based IIS tends to show improved performance in less complex datasets as its effectiveness largely relies on precise classification outcomes.

\textbf{Sensitivity to Hyper-parameters.} The optimal maximum pseudo bag number $M_{\text{max}}$ varies across datasets and heavily relies on the number of positive instances present. For datasets containing larger tumor regions within bags, a higher $M_{\text{max}}$ is recommended. Conversely, datasets with fewer positive instances per bag benefit from a smaller $M_{\text{max}}$ to avoid unnecessary complexity. Our model exhibits insensitivity to hyper-parameters used in Shapley value approximation. Nevertheless, it is advisable not to set $\mu$ and $\tau$ too small.

\textbf{Progressive Strategies.} A progressive increase in the number of pseudo bags is effective for challenging datasets or those with only a limited number of positive instances in each bag. While this approach is less appealing for datasets with substantial tumor regions. Conversely, progressive initialization represents a significant improvement across various datasets, especially on more challenging ones.

\section{Conclusion}
In this study, we tackle attention-related challenges within multiple-instance learning for whole-slide image classification, focusing on the issues of extreme attention distribution and misidentification of positive instances. To overcome these challenges, we introduce the accelerated Shapley value to quantify the contribution of each instance, pioneering the estimation of IIS for the first time. Compared with traditional pseudo bag assignment strategies like random splitting, our approach provides a more logical allocation of pseudo bags, effectively mitigating the mislabeling issue to some extent. Furthermore, we present a progressive pseudo bag augmented MIL framework that incorporates Shapley value-based IIS and utilizes the expectation-maximization algorithm. This framework systematically enhances pseudo bag augmentation, significantly boosting the efficacy of MIL. Extensive experiments on four publicly available datasets demonstrate that our methodology surpasses existing state-of-the-art techniques. Additionally, the Shapley value-based IIS offers valuable class-wise interpretability. Superior performance and interpretation help pathologists make more comprehensive and efficient diagnoses in clinical practice.

In future research, we aim to investigate alternative metrics for precise and efficient estimation of IIS, striving to enhance the robustness and versatility of our proposed framework.

\bibliographystyle{IEEEtran}
\bibliography{main}

\begin{thebibliography}{10}
\providecommand{\url}[1]{#1}
\csname url@samestyle\endcsname
\providecommand{\newblock}{\relax}
\providecommand{\bibinfo}[2]{#2}
\providecommand{\BIBentrySTDinterwordspacing}{\spaceskip=0pt\relax}
\providecommand{\BIBentryALTinterwordstretchfactor}{4}
\providecommand{\BIBentryALTinterwordspacing}{\spaceskip=\fontdimen2\font plus
\BIBentryALTinterwordstretchfactor\fontdimen3\font minus
  \fontdimen4\font\relax}
\providecommand{\BIBforeignlanguage}[2]{{%
\expandafter\ifx\csname l@#1\endcsname\relax
\typeout{** WARNING: IEEEtran.bst: No hyphenation pattern has been}%
\typeout{** loaded for the language `#1'. Using the pattern for}%
\typeout{** the default language instead.}%
\else
\language=\csname l@#1\endcsname
\fi
#2}}
\providecommand{\BIBdecl}{\relax}
\BIBdecl

\bibitem{ba2022assessment}
W.~Ba, S.~Wang, M.~Shang, Z.~Zhang, H.~Wu, C.~Yu, R.~Xing, W.~Wang, L.~Wang,
  C.~Liu \emph{et~al.}, ``Assessment of deep learning assistance for the
  pathological diagnosis of gastric cancer,'' \emph{Modern Pathology}, vol.~35,
  no.~9, pp. 1262--1268, 2022.

\bibitem{schapiro2017histocat}
D.~Schapiro, H.~W. Jackson, S.~Raghuraman, J.~R. Fischer, V.~R. Zanotelli,
  D.~Schulz, C.~Giesen, R.~Catena, Z.~Varga, and B.~Bodenmiller, ``histocat:
  analysis of cell phenotypes and interactions in multiplex image cytometry
  data,'' \emph{Nature methods}, vol.~14, no.~9, pp. 873--876, 2017.

\bibitem{bera2019artificial}
K.~Bera, K.~A. Schalper, D.~L. Rimm, V.~Velcheti, and A.~Madabhushi,
  ``Artificial intelligence in digital pathology—new tools for diagnosis and
  precision oncology,'' \emph{Nature reviews Clinical oncology}, vol.~16,
  no.~11, pp. 703--715, 2019.

\bibitem{niazi2019digital}
M.~K.~K. Niazi, A.~V. Parwani, and M.~N. Gurcan, ``Digital pathology and
  artificial intelligence,'' \emph{The lancet oncology}, vol.~20, no.~5, pp.
  e253--e261, 2019.

\bibitem{moen2019deep}
E.~Moen, D.~Bannon, T.~Kudo, W.~Graf, M.~Covert, and D.~Van~Valen, ``Deep
  learning for cellular image analysis,'' \emph{Nature methods}, vol.~16,
  no.~12, pp. 1233--1246, 2019.

\bibitem{8756037}
F.~Mahmood, D.~Borders, R.~J. Chen, G.~N. Mckay, K.~J. Salimian, A.~Baras, and
  N.~J. Durr, ``Deep adversarial training for multi-organ nuclei segmentation
  in histopathology images,'' \emph{IEEE Transactions on Medical Imaging},
  vol.~39, no.~11, pp. 3257--3267, 2020.

\bibitem{lin2022pdbl}
J.~Lin, G.~Han, X.~Pan, Z.~Liu, H.~Chen, D.~Li, X.~Jia, Z.~Shi, Z.~Wang, Y.~Cui
  \emph{et~al.}, ``Pdbl: Improving histopathological tissue classification with
  plug-and-play pyramidal deep-broad learning,'' \emph{IEEE Transactions on
  Medical Imaging}, vol.~41, no.~9, pp. 2252--2262, 2022.

\bibitem{yan2023unpaired}
R.~Yan, Q.~He, Y.~Liu, P.~Ye, L.~Zhu, S.~Shi, J.~Gou, Y.~He, T.~Guan, and
  G.~Zhou, ``Unpaired virtual histological staining using prior-guided
  generative adversarial networks,'' \emph{Computerized Medical Imaging and
  Graphics}, vol. 105, p. 102185, 2023.

\bibitem{pinheiro2015image}
P.~O. Pinheiro and R.~Collobert, ``From image-level to pixel-level labeling
  with convolutional networks,'' in \emph{Proceedings of the IEEE conference on
  computer vision and pattern recognition}, 2015, pp. 1713--1721.

\bibitem{bejnordi2017diagnostic}
B.~E. Bejnordi, M.~Veta, P.~J. Van~Diest, B.~Van~Ginneken, N.~Karssemeijer,
  G.~Litjens, J.~A. Van Der~Laak, M.~Hermsen, Q.~F. Manson, M.~Balkenhol
  \emph{et~al.}, ``Diagnostic assessment of deep learning algorithms for
  detection of lymph node metastases in women with breast cancer,''
  \emph{Jama}, vol. 318, no.~22, pp. 2199--2210, 2017.

\bibitem{campanella2019clinical}
G.~Campanella, M.~G. Hanna, L.~Geneslaw, A.~Miraflor, V.~Werneck Krauss~Silva,
  K.~J. Busam, E.~Brogi, V.~E. Reuter, D.~S. Klimstra, and T.~J. Fuchs,
  ``Clinical-grade computational pathology using weakly supervised deep
  learning on whole slide images,'' \emph{Nature medicine}, vol.~25, no.~8, pp.
  1301--1309, 2019.

\bibitem{lee2022derivation}
Y.~Lee, J.~H. Park, S.~Oh, K.~Shin, J.~Sun, M.~Jung, C.~Lee, H.~Kim, J.-H.
  Chung, K.~C. Moon \emph{et~al.}, ``Derivation of prognostic contextual
  histopathological features from whole-slide images of tumours via graph deep
  learning,'' \emph{Nature Biomedical Engineering}, pp. 1--15, 2022.

\bibitem{pati2022hierarchical}
P.~Pati, G.~Jaume, A.~Foncubierta-Rodriguez, F.~Feroce, A.~M. Anniciello,
  G.~Scognamiglio, N.~Brancati, M.~Fiche, E.~Dubruc, D.~Riccio \emph{et~al.},
  ``Hierarchical graph representations in digital pathology,'' \emph{Medical
  image analysis}, vol.~75, p. 102264, 2022.

\bibitem{li2023deeptree}
J.~Li, J.~Cheng, L.~Meng, H.~Yan, Y.~He, H.~Shi, T.~Guan, and A.~Han,
  ``Deeptree: Pathological image classification through imitating tree-like
  strategies of pathologists,'' \emph{IEEE Transactions on Medical Imaging},
  2023.

\bibitem{maron1997framework}
O.~Maron and T.~Lozano-P{\'e}rez, ``A framework for multiple-instance
  learning,'' \emph{Advances in neural information processing systems},
  vol.~10, 1997.

\bibitem{feng2017deep}
J.~Feng and Z.-H. Zhou, ``Deep miml network,'' in \emph{Proceedings of the AAAI
  conference on artificial intelligence}, vol.~31, 2017.

\bibitem{zhu2017deep}
W.~Zhu, Q.~Lou, Y.~S. Vang, and X.~Xie, ``Deep multi-instance networks with
  sparse label assignment for whole mammogram classification,'' in
  \emph{International conference on medical image computing and
  computer-assisted intervention}.\hskip 1em plus 0.5em minus 0.4em\relax
  Springer, 2017, pp. 603--611.

\bibitem{li2021multi}
J.~Li, W.~Li, A.~Sisk, H.~Ye, W.~D. Wallace, W.~Speier, and C.~W. Arnold, ``A
  multi-resolution model for histopathology image classification and
  localization with multiple instance learning,'' \emph{Computers in biology
  and medicine}, vol. 131, p. 104253, 2021.

\bibitem{wang2022weakly}
X.~Wang, H.~Chen, C.~Gan, H.~Lin, Q.~Dou, Q.~Huang, M.~Cai, and P.-A. Heng,
  ``Weakly supervised learning for whole slide lung cancer image
  classification,'' in \emph{Medical imaging with deep learning}, 2022.

\bibitem{pmlr-v143-sharma21a}
Y.~Sharma, A.~Shrivastava, L.~Ehsan, C.~A. Moskaluk, S.~Syed, and D.~Brown,
  ``Cluster-to-conquer: A framework for end-to-end multi-instance learning for
  whole slide image classification,'' in \emph{Proceedings of the Fourth
  Conference on Medical Imaging with Deep Learning}, ser. Proceedings of
  Machine Learning Research, vol. 143.\hskip 1em plus 0.5em minus 0.4em\relax
  PMLR, 07--09 Jul 2021, pp. 682--698.

\bibitem{chen2022scaling}
R.~J. Chen, C.~Chen, Y.~Li, T.~Y. Chen, A.~D. Trister, R.~G. Krishnan, and
  F.~Mahmood, ``Scaling vision transformers to gigapixel images via
  hierarchical self-supervised learning,'' in \emph{Proceedings of the IEEE/CVF
  Conference on Computer Vision and Pattern Recognition}, 2022, pp.
  16\,144--16\,155.

\bibitem{zhu2023accurate}
L.~Zhu, H.~Shi, H.~Wei, C.~Wang, S.~Shi, F.~Zhang, R.~Yan, Y.~Liu, T.~He,
  L.~Wang \emph{et~al.}, ``An accurate prediction of the origin for bone
  metastatic cancer using deep learning on digital pathological images,''
  \emph{EBioMedicine}, vol.~87, 2023.

\bibitem{lin2023}
T.~Lin, Z.~Yu, H.~Hu, Y.~Xu, and C.-W. Chen, ``Interventional bag
  multi-instance learning on whole-slide pathological images,'' in
  \emph{Proceedings of the IEEE/CVF Conference on Computer Vision and Pattern
  Recognition}, 2023, pp. 19\,830--19\,839.

\bibitem{ilse2018attention}
M.~Ilse, J.~Tomczak, and M.~Welling, ``Attention-based deep multiple instance
  learning,'' in \emph{International conference on machine learning}.\hskip 1em
  plus 0.5em minus 0.4em\relax PMLR, 2018, pp. 2127--2136.

\bibitem{lu2021data}
M.~Y. Lu, D.~F. Williamson, T.~Y. Chen, R.~J. Chen, M.~Barbieri, and
  F.~Mahmood, ``Data-efficient and weakly supervised computational pathology on
  whole-slide images,'' \emph{Nature biomedical engineering}, vol.~5, no.~6,
  pp. 555--570, 2021.

\bibitem{li2021dual}
B.~Li, Y.~Li, and K.~W. Eliceiri, ``Dual-stream multiple instance learning
  network for whole slide image classification with self-supervised contrastive
  learning,'' in \emph{Proceedings of the IEEE/CVF conference on computer
  vision and pattern recognition}, 2021, pp. 14\,318--14\,328.

\bibitem{yufei2022bayes}
C.~Yufei, Z.~Liu, X.~Liu, X.~Liu, C.~Wang, T.-W. Kuo, C.~J. Xue, and A.~B.
  Chan, ``Bayes-mil: A new probabilistic perspective on attention-based
  multiple instance learning for whole slide images,'' in \emph{The Eleventh
  International Conference on Learning Representations}, 2022.

\bibitem{zhang2023attention}
Y.~Zhang, H.~Li, Y.~Sun, S.~Zheng, C.~Zhu, and L.~Yang, ``Attention-challenging
  multiple instance learning for whole slide image classification,''
  \emph{arXiv preprint arXiv:2311.07125}, 2023.

\bibitem{qu2022bi}
L.~Qu, M.~Wang, Z.~Song \emph{et~al.}, ``Bi-directional weakly supervised
  knowledge distillation for whole slide image classification,'' \emph{Advances
  in Neural Information Processing Systems}, vol.~35, pp. 15\,368--15\,381,
  2022.

\bibitem{yu2023bayesian}
J.-G. Yu, Z.~Wu, Y.~Ming, S.~Deng, Q.~Wu, Z.~Xiong, T.~Yu, G.-S. Xia, Q.~Jiang,
  and Y.~Li, ``Bayesian collaborative learning for whole-slide image
  classification,'' \emph{IEEE Transactions on Medical Imaging}, 2023.

\bibitem{li2023task}
H.~Li, C.~Zhu, Y.~Zhang, Y.~Sun, Z.~Shui, W.~Kuang, S.~Zheng, and L.~Yang,
  ``Task-specific fine-tuning via variational information bottleneck for
  weakly-supervised pathology whole slide image classification,'' in
  \emph{Proceedings of the IEEE/CVF Conference on Computer Vision and Pattern
  Recognition}, 2023, pp. 7454--7463.

\bibitem{shapley1953value}
L.~S. Shapley \emph{et~al.}, ``A value for n-person games,'' 1953.

\bibitem{messalas2019model}
A.~Messalas, Y.~Kanellopoulos, and C.~Makris, ``Model-agnostic interpretability
  with shapley values,'' in \emph{2019 10th International Conference on
  Information, Intelligence, Systems and Applications (IISA)}.\hskip 1em plus
  0.5em minus 0.4em\relax IEEE, 2019, pp. 1--7.

\bibitem{tang2021data}
S.~Tang, A.~Ghorbani, R.~Yamashita, S.~Rehman, J.~A. Dunnmon, J.~Zou, and D.~L.
  Rubin, ``Data valuation for medical imaging using shapley value and
  application to a large-scale chest x-ray dataset,'' \emph{Scientific
  reports}, vol.~11, no.~1, p. 8366, 2021.

\bibitem{wang2022scl}
X.~Wang, J.~Xiang, J.~Zhang, S.~Yang, Z.~Yang, M.-H. Wang, J.~Zhang, W.~Yang,
  J.~Huang, and X.~Han, ``Scl-wc: Cross-slide contrastive learning for
  weakly-supervised whole-slide image classification,'' \emph{Advances in
  neural information processing systems}, vol.~35, pp. 18\,009--18\,021, 2022.

\bibitem{shao2021transmil}
Z.~Shao, H.~Bian, Y.~Chen, Y.~Wang, J.~Zhang, X.~Ji \emph{et~al.}, ``Transmil:
  Transformer based correlated multiple instance learning for whole slide image
  classification,'' \emph{Advances in neural information processing systems},
  vol.~34, pp. 2136--2147, 2021.

\bibitem{shao2021weakly}
W.~Shao, T.~Wang, Z.~Huang, Z.~Han, J.~Zhang, and K.~Huang, ``Weakly supervised
  deep ordinal cox model for survival prediction from whole-slide pathological
  images,'' \emph{IEEE Transactions on Medical Imaging}, vol.~40, no.~12, pp.
  3739--3747, 2021.

\bibitem{yang2023protodiv}
R.~Yang, P.~Liu, and L.~Ji, ``Protodiv: Prototype-guided division of consistent
  pseudo-bags for whole-slide image classification,'' \emph{arXiv preprint
  arXiv:2304.06652}, 2023.

\bibitem{zhang2022dtfd}
H.~Zhang, Y.~Meng, Y.~Zhao, Y.~Qiao, X.~Yang, S.~E. Coupland, and Y.~Zheng,
  ``Dtfd-mil: Double-tier feature distillation multiple instance learning for
  histopathology whole slide image classification,'' in \emph{Proceedings of
  the IEEE/CVF Conference on Computer Vision and Pattern Recognition}, 2022,
  pp. 18\,802--18\,812.

\bibitem{liu2024pseudo}
P.~Liu, L.~Ji, X.~Zhang, and F.~Ye, ``Pseudo-bag mixup augmentation for
  multiple instance learning-based whole slide image classification,''
  \emph{IEEE Transactions on Medical Imaging}, 2024.

\bibitem{javed2022additive}
S.~A. Javed, D.~Juyal, H.~Padigela, A.~Taylor-Weiner, L.~Yu, and A.~Prakash,
  ``Additive mil: intrinsically interpretable multiple instance learning for
  pathology,'' \emph{Advances in Neural Information Processing Systems},
  vol.~35, pp. 20\,689--20\,702, 2022.

\bibitem{strumbelj2010efficient}
E.~Strumbelj and I.~Kononenko, ``An efficient explanation of individual
  classifications using game theory,'' \emph{The Journal of Machine Learning
  Research}, vol.~11, pp. 1--18, 2010.

\bibitem{lundberg2017unified}
S.~M. Lundberg and S.-I. Lee, ``A unified approach to interpreting model
  predictions,'' \emph{Advances in neural information processing systems},
  vol.~30, 2017.

\bibitem{chen2018shapley}
J.~Chen, L.~Song, M.~J. Wainwright, and M.~I. Jordan, ``L-shapley and
  c-shapley: Efficient model interpretation for structured data,'' \emph{arXiv
  preprint arXiv:1808.02610}, 2018.

\bibitem{ancona2019explaining}
M.~Ancona, C.~Oztireli, and M.~Gross, ``Explaining deep neural networks with a
  polynomial time algorithm for shapley value approximation,'' in
  \emph{International Conference on Machine Learning}.\hskip 1em plus 0.5em
  minus 0.4em\relax PMLR, 2019, pp. 272--281.

\bibitem{dempster1977maximum}
A.~P. Dempster, N.~M. Laird, and D.~B. Rubin, ``Maximum likelihood from
  incomplete data via the em algorithm,'' \emph{Journal of the royal
  statistical society: series B (methodological)}, vol.~39, no.~1, pp. 1--22,
  1977.

\bibitem{brancati2022bracs}
N.~Brancati, A.~M. Anniciello, P.~Pati, D.~Riccio, G.~Scognamiglio, G.~Jaume,
  G.~De~Pietro, M.~Di~Bonito, A.~Foncubierta, G.~Botti \emph{et~al.}, ``Bracs:
  A dataset for breast carcinoma subtyping in h\&e histology images,''
  \emph{Database}, vol. 2022, p. baac093, 2022.

\bibitem{he2016deep}
K.~He, X.~Zhang, S.~Ren, and J.~Sun, ``Deep residual learning for image
  recognition,'' in \emph{Proceedings of the IEEE conference on computer vision
  and pattern recognition}, 2016, pp. 770--778.

\end{thebibliography}

\end{document}